\documentclass[runningheads,a4paper]{llncs}

  \usepackage{amssymb}
  \usepackage{amsmath}
  \usepackage{graphicx}
  \graphicspath{{images/}}
  \usepackage{subfigure}

  \usepackage{booktabs}
  \usepackage{multirow}
  \usepackage{array}
  \usepackage{tabularx}
  \newcolumntype{x}[1]{>{\centering\arraybackslash\hspace{0pt}}p{#1}}
  
  \usepackage[inline]{enumitem}
  \usepackage{bbm}
  
  \usepackage{url}
  \urldef{\mailsc}\path|hliao6@cs.rochester.edu|

  \usepackage[misc]{ifsym}
  
\begin{document}
  
  \mainmatter  
  
  \title{Adversarial Sparse-View CBCT \\ Artifact Reduction}
  
  
  %
  %
  \author{Haofu Liao\textsuperscript{1 (\Letter)}, %
  Zhimin Huo\textsuperscript{2},
  William J. Sehnert\textsuperscript{2},
  Shaohua Kevin Zhou\textsuperscript{3},\\
  \and Jiebo Luo\textsuperscript{1}}
  
  \authorrunning{H. Liao et al}
  
  \institute{\textsuperscript{1} Department of Computer Science, University of Rochester, Rochester, USA\\
  \mailsc\\
  \textsuperscript{2} Carestream Health Inc., Rochester, USA\\
  \textsuperscript{3} Institute of Computing Technology, Chinese Academy of Sciences, Beijing, China}
  
  %
  %
  
  \toctitle{Adversarial CBCT Streak Artifact Reduction}
  \tocauthor{H. Liao et al.}
  \maketitle

  \begin{abstract}
  
  We present an effective post-processing method to reduce the artifacts from sparsely reconstructed cone-beam CT (CBCT) images. The proposed method is based on the state-of-the-art, image-to-image generative models with a perceptual loss as regulation. Unlike the traditional CT artifact-reduction approaches, our method is trained in an adversarial fashion that yields more perceptually realistic outputs while preserving the anatomical structures. To address the streak artifacts that are inherently local and appear across various scales, we further propose a novel discriminator architecture based on feature pyramid networks and a differentially modulated focus map to induce the adversarial training. Our experimental results show that the proposed method can greatly correct the cone-beam artifacts from clinical CBCT images reconstructed using 1/3 projections, and outperforms strong baseline methods both quantitatively and qualitatively.
  
  \end{abstract}
  
  \section{Introduction}
  
  Cone-beam computed tomography (CBCT) is a variant type of computed tomography (CT). Compared with conventional CT,
  CBCT usually 
  has shorter examination time, resulting in fewer motion artifacts and better X-ray tube efficiency. One way to further shorten the acquisition time and enhance the healthcare experience is to take fewer X-ray measurements during each CBCT scan. However, due to the ``cone-beam'' projection geometry, CBCT images typically contain more pronounced streak artifacts than CT images and this is even worse when fewer X-ray projections are used during the CBCT reconstruction \cite{bian2010evaluation}.
  
  
  
  A number of approaches have been proposed to address the artifacts \cite{zhang2007reducing,ning2004x}
  that are commonly encountered in CBCT images.
  However, to our best knowledge, no scheme has been proposed to correct the cone-beam artifacts introduced by sparse-view CBCT reconstruction in a post-processing step. 
  Instead of reducing artifacts from the CBCT images directly, many other systems  \cite{xia2016optimization,zhang2016artifact} propose to introduce better sparse-view reconstruction methods that yield less artifacts. Although encouraging improvements have been made, the image quality from the current solutions are still not satisfactory when only a small number of views are used. This work attempts to fill this gap by refining the sparsely reconstructed CBCT images through a novel cone-beam artifact reduction method.
  
  In relation to this study, there are many works that leverage deep neural networks (DNNs) for low-dose CT (LDCT) denoising.
  \cite{chen2017low2} used a residual encoder-decoder architecture to reduce the noise from LDCT images, and achieved superior performance over traditional approaches.
  More recently, \cite{wolterink2017generative} introduced generative adversarial networks (GANs) \cite{goodfellow2014generative} into their architecture to obtain more realistic outputs, and this work was further improved by \cite{yang2017low} where a combination of perceptual loss \cite{johnson2016perceptual} and adversarial loss was used.

  Similarly, this work also proposes to use DNNs for sparse-view CBCT artifact reduction. 
  We train an image-to-image generative model with perceptual loss to obtain outputs that are perceptually close to the dense-view CBCT images. To address the artifacts at various levels, we further contribute to the literature with a novel discriminator architecture based on feature pyramid networks (FPN) \cite{lin2016feature} and a differentially modulated focus map so that the adversarial training is biased to the artifacts at multiple scales. The proposed approach is evaluated on clinical CBCT images. Experimental results demonstrate that our method outperforms strong baseline methods both qualitatively and quantitatively.
  
  \begin{figure}[t]
    \includegraphics[width=0.9\textwidth]{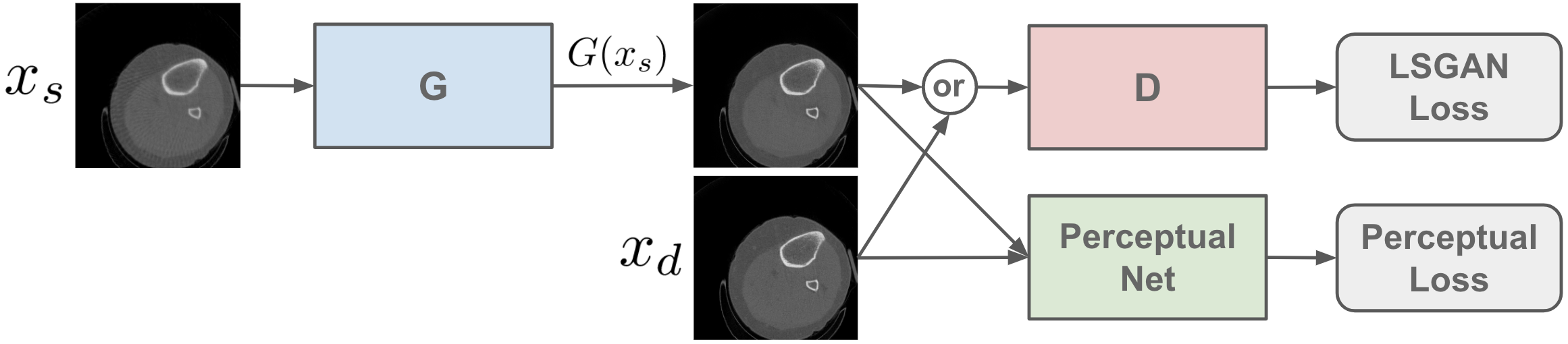}
    \centering
    \caption{The overall architecture of the proposed method.}
    \label{fig:architecture}
  \end{figure}
  
  \section{Methods}
  
  
  Let $x_{s}$ be a sparse-view CBCT image, which is reconstructed from a sparse set or low number of projections (or views), and $x_{d}$ be its dense-view counterpart, which is reconstructed from a dense set or a high number of projections (or views). The proposed method is formulated under an image-to-image generative model as illustrated in Fig. \ref{fig:architecture} where we train a generator that transforms $x_{s}$ to an ideally artifact-free image that looks like $x_{d}$. The discriminator is used for the adversarial training, and the perceptual network is included for additional perceptual and structural regularization. We use LSGAN \cite{mao2016least} against a regular GAN to achieve more stable adversarial learning. The adversarial objective functions for the proposed model can be written as
  \begin{align}
    \min_{D} \mathcal{L}_A(D;G, \Lambda) &= \mathbb{E}_{\mathbf{X}_{d}}[\lVert \Lambda \odot (D(x_d) - \mathbf{1}) \rVert^2] + \mathbb{E}_{\mathbf{X}_{s}}[\lVert \Lambda \odot D(G(x_s)) \rVert^2] \label{eq: D}, \\
    \min_{G} \mathcal{L}_A(G; D, \Lambda) &= \mathbb{E}_{\mathbf{X}_{s}}[\lVert \Lambda \odot (D(G(x_s)) - \mathbf{1}) \rVert^2] \label{eq: G},
  \end{align}
  where $\Lambda$ is a focus map detailed in Sec. \ref{sec: focal_loss}. Here, we apply a PatchGAN-like \cite{isola2016image} design to the discriminator so that the realness is patch based and the output is a score map. The generator $G$ and discriminator $D$ are trained in an adversarial fashion. $D$ distinguishes between $x_{d}$ and the generated CBCT image $G(x_s)$ (Eq. \ref{eq: D}), while $G$ generates CBCT image samples as ``real'' as possible so that $D$ cannot tell if they are dense-view CBCT images or generated by $G$ (Eq. \ref{eq: G}).
  
  Training with the adversarial loss, alone, usually introduces additional artifacts, and previous works often use MSE loss to induce the learning \cite{wolterink2017generative}. However, as shown by \cite{yang2017low}, MSE loss does not handle streak artifacts very well. Therefore, we adopt the choice of \cite{yang2017low} by using a perceptual loss to induce the learning and give more realistic outputs. Let $\phi^{(i)}(\cdot)$ denote the feature maps extracted by the $i$-th layer of the perceptual network $\phi$ and $N_i$ denote the number of elements in $\phi^{(i)}(\cdot)$, the perceptual loss can be computed by
  \begin{equation}
    \mathcal{L}_{P} =
    \frac{1}{N_i}
    \lVert \phi^{(i)}(x_{d}) - \phi^{(i)}(G(x_{s})) \rVert_1.
  \end{equation}
  In this work, the perceptual network $\phi$ is a pretrained VGG16 net \cite{simonyan2014very} and we empirically find that $i=8$ works well.
  
  \subsection{Network Structure}
  
  \begin{figure}[t]
    \includegraphics[width=\textwidth]{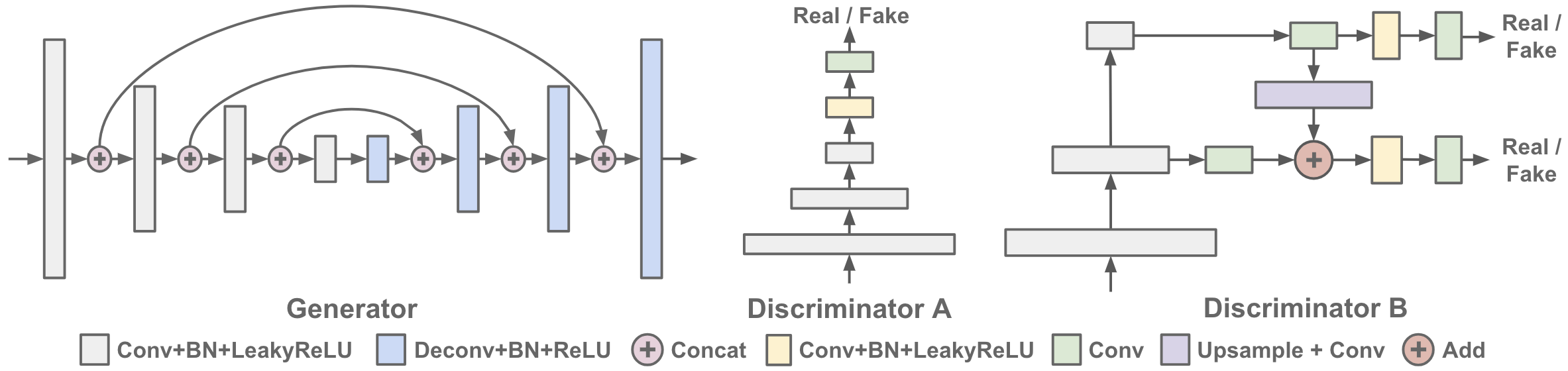}
    \centering
    \caption{Detailed network structure of the generator and discriminator.}
    \label{fig:g_d}
  \end{figure}
  
  The generator is based on an encoder-decoder architecture \cite{isola2016image}.
  As shown in Fig. \ref{fig:g_d}, the generator has four encoding blocks (in gray) and four decoding blocks (in blue). Each encoding block contains a convolutional layer followed by a batch normalization layer and a leaky ReLU layer. Similarly, each decoding block contains a deconvolutional layer followed by a batch normalization layer and a ReLU layer. Both the convolutional and deconvolutional layers have a $4 \times 4$ kernel with a stride of $2$ so that they can downsample and upsample the outputs, respectively. Outputs from the encoding blocks are shuttled to the corresponding decoding blocks using skip connections. 
  This design allows the low-level context information from the encoding blocks to be used directly together with the decoded high-level information during generation. 
  
  A typical discriminator (Fig. \ref{fig:g_d} Discriminator A) usually contains a set of encoding blocks followed by a classifier to determine the input's realness. In this case, the discrimination is performed at a fixed granularity that is fine when the task is a generative task such as style transfer or image translation, or there is a systematic error to be corrected such as JPEG decompression or super-resolution. For sparse-view CBCT images, the artifacts appear randomly with different scales. To capture such a variation of artifacts, we propose a discriminator that handles the adversarial training at different granularities.

  The core idea is to create a feature pyramid and perform discrimination at multiple scales. As illustrated in Fig. \ref{fig:g_d} Discriminator B, the network uses two outputs and makes decisions based on different levels of semantic feature maps. We adapt the design from FPN \cite{lin2016feature} so that the feature pyramid has strong semantics at all scales. Specifically, we first use three encoding blocks to extract features at different levels. 
  Next, we use an upsample block (in purple) to incorporate the stronger semantic features from the top layer into the outputs of the middle layer. The upsample block consists of a 
  unsampling layer and a $3 \times 3$ convolutional layer (to smooth the outputs). Because the feature maps from the encoding blocks have different channel sizes, we place a lateral block (in green, essentially a $1 \times 1$ convolutional layer) after each encoding block to match this channel difference. In the end, there are two classifiers to make joint decisions on the semantics at different scales. Each classifier contains two blocks. The first block (in yellow) has the same layers as an encoding block, except that the convolutional layer has a $3 \times 3$ kernel with a stride of $1$. The second block (in green) is simply a $1 \times 1$ convolutional layer with stride $1$. Let $D_1(x)$ and $D_2(x)$ denote the outputs from the two classifiers, then the new adversarial loss can be given by
  $\min_{D} \mathcal{L}_A(D;G, \Lambda_1, \Lambda_2) = \sum_{i=1}^2 \mathcal{L}_A(D_i;G, \Lambda_i)$
  and
  $\min_{G} \mathcal{L}_A(G; D, \Lambda_1, \Lambda_2) = \sum_{i=1}^2 \mathcal{L}_A(G;D_i, \Lambda_i)$.
  We also experimented with deeper discriminators with more classifiers for richer feature semantics, but found that they contribute only minor improvements over the current setting.
  
  \subsection{Focus Map} \label{sec: focal_loss}

  \begin{figure}[t]
    \centering
    \subfigure[$x_{d}$]{
    \includegraphics[width=.2\textwidth]{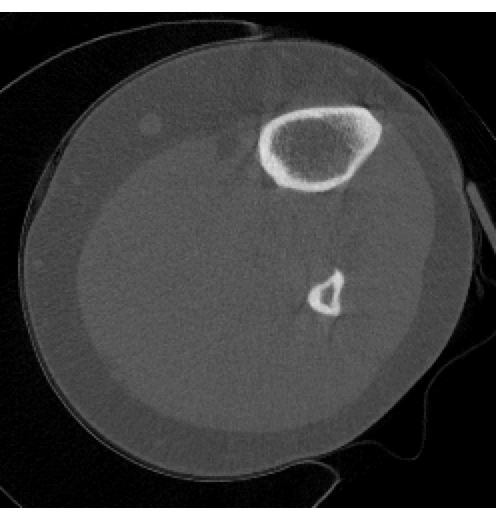}
    }
    \hspace*{-1.1em}
    \subfigure[$G(x_{s})$]{
    \includegraphics[width=.2\textwidth]{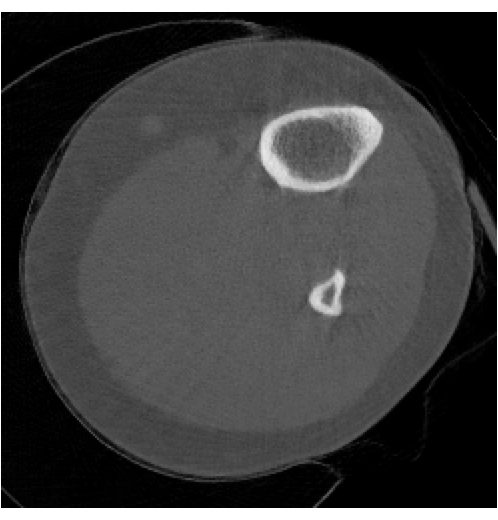}
    }
    \hspace*{-1.1em}
    \subfigure[$D(G(x_{s}))$]{
    \includegraphics[width=.236\textwidth]{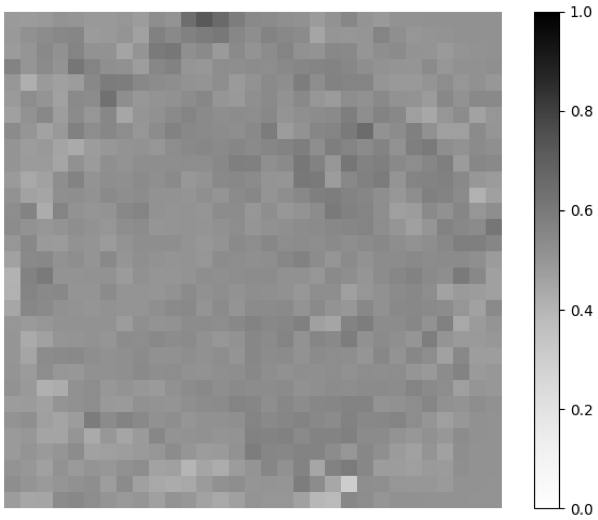}
    }
    \hspace*{-1.0em}
    \subfigure[$\Lambda$]{
    \includegraphics[width=.236\textwidth]{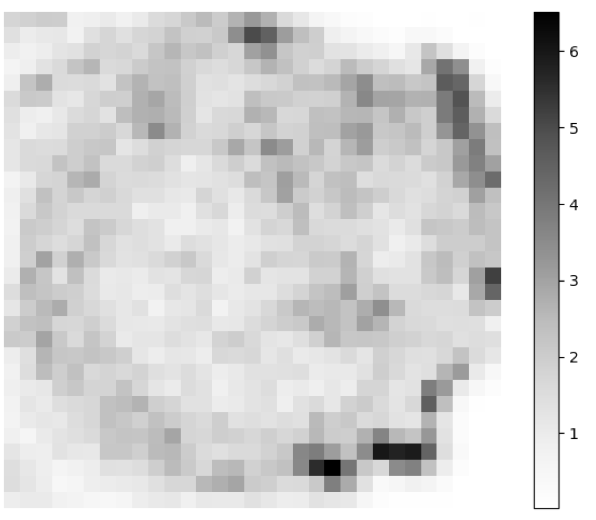}
    }
  
    \caption{
      Saturated (c) score map $D(G(x_{s}))$ and (d) focus map $\Lambda$ computed between (a) dense-view CBCT image $x_{d}$ and (b) generated CBCT image $G(x_{s})$.
    }
    \label{fig:feature_difference}
  \end{figure}
  
  When an image from the generator looks mostly ``real'' (Fig \ref{fig:feature_difference} (b)),  the score map (Fig \ref{fig:feature_difference} (c)) output by the discriminator will be overwhelmed by borderline scores (those values close to 0.5). This saturates the adversarial training as borderline scores make little contribution to the weight update of the discriminator. To address this problem, we propose to introduce a modulation factor to the adversarial loss so that the borderline scores are down-weighted during training. Observing that when a generated region is visually close to the corresponding region of a dense-view image (Fig \ref{fig:feature_difference} (a)), it is more likely to be ``real'' and causes the discriminator to give a borderline score. Therefore, we use a feature difference map (Fig \ref{fig:feature_difference} (d)) to perform this modulation.
  
  Let $\phi_{m,n}^{(j)}(\cdot)$ denote the $(m,n)$-th feature vector of $\phi^{(j)}(\cdot)$, then the $(m,n)$-th element of the feature difference map $\Lambda$ between $x_{d}$ and $G(x_{s})$ is defined as
  \begin{equation}
    \lambda_{m,n} = \frac{1}{Z_j} \lVert \phi_{m,n}^{(j)}(x_{d}) - \phi_{m,n}^{(j)}(G(x_{s})) \rVert,
  \end{equation}
  where $Z_j$ is a normalization term given by
  \begin{equation}
    Z_j = \frac{1}{N_j} \sum_{m,n} \lVert \phi_{m,n}^{(j)}(x_{d}) - \phi_{m,n}^{(j)}(G(x_{s})) \rVert.
  \end{equation}
  We use the same perceptual network $\phi$ as the one used for computing the perceptual loss, and $j$ is chosen to match the resolution of $D_1(x)$ and $D_2(x)$. For the VGG16 net, we use $j=16$ for $\Lambda_1$ and $j=9$ for $\Lambda_2$.
  
  \section{Experiments}
  
  
  
  \noindent \textbf{Datasets} The CBCT images were obtained by a multi-source CBCT scanner dedicated for lower extremities. In total, knee images from 27 subjects are under investigation. Each subject is associated with a sparse-view image and a dense-view image that are reconstructed 
  using 67 and 200 projection views, respectively. Each image is processed, slice by slice, along the sagittal direction where the streak artifacts are most pronounced. During the training, the inputs to the models are $256 \times 256$ patches that randomly cropped from the slices. \vspace{0.5em}

  \noindent \textbf{Models} Three variants of the proposed methods as well as two other baseline methods are compared:
  \begin{enumerate*}[label=(\roman*)]
    \item Baseline-MSE: a similar approach to \cite{wolterink2017generative} by combining MSE loss with GAN. 3D UNet\footnote{Identical to the 2D UNet used in this work with all the 2D convolutional and deconvolutional layers replaced by their 3D counterparts.} and LSGAN is used for fair comparison;
    
    \item Baseline-Perceptual: a similar approach to \cite{yang2017low} by combining perceptual loss with GAN. It is also based on our UNet and LSGAN infrastructure for fair comparison;
    
    \item Ours-FPN: our method using FPN as the discriminator and setting $\Lambda_1 = \Lambda_2 = \mathbf{1}$;
    
    \item Ours-Focus: our method using focus map and conventional discriminator (Fig. \ref{fig:g_d} Discriminator A);
    \item Ours-Focus+FPN: our method using focus map as well as the FPN discriminator.
  \end{enumerate*}
  We train all the models using Adam optimization with the learning rate $lr=10^{-4}$ and $\beta_1=0.5$. We use $\lambda_{a} = 1.0$, $ \lambda_{m} = 100$, and $\lambda_{p}=10$ to control the weights between the adversarial loss, the MSE loss, and the perceptual loss. The values are chosen empirically and are the same for all models (if applicable). All the models are trained for 50 epochs with 5-fold cross-validation. We perform all the experiments on an Nvidia GeForce GTX 1070 GPU. During testing, the average processing time on $384 \times \times 384 \times 417$ CBCT volumes for the 2D UNet (generator of model (ii)-(v)) is 16.05 seconds, and for the 3D UNet (generator of model (i)) is 22.70 seconds. \vspace{0.5em}

  \begin{figure}[!t]
    \centering
    \includegraphics[width=.8\textwidth]{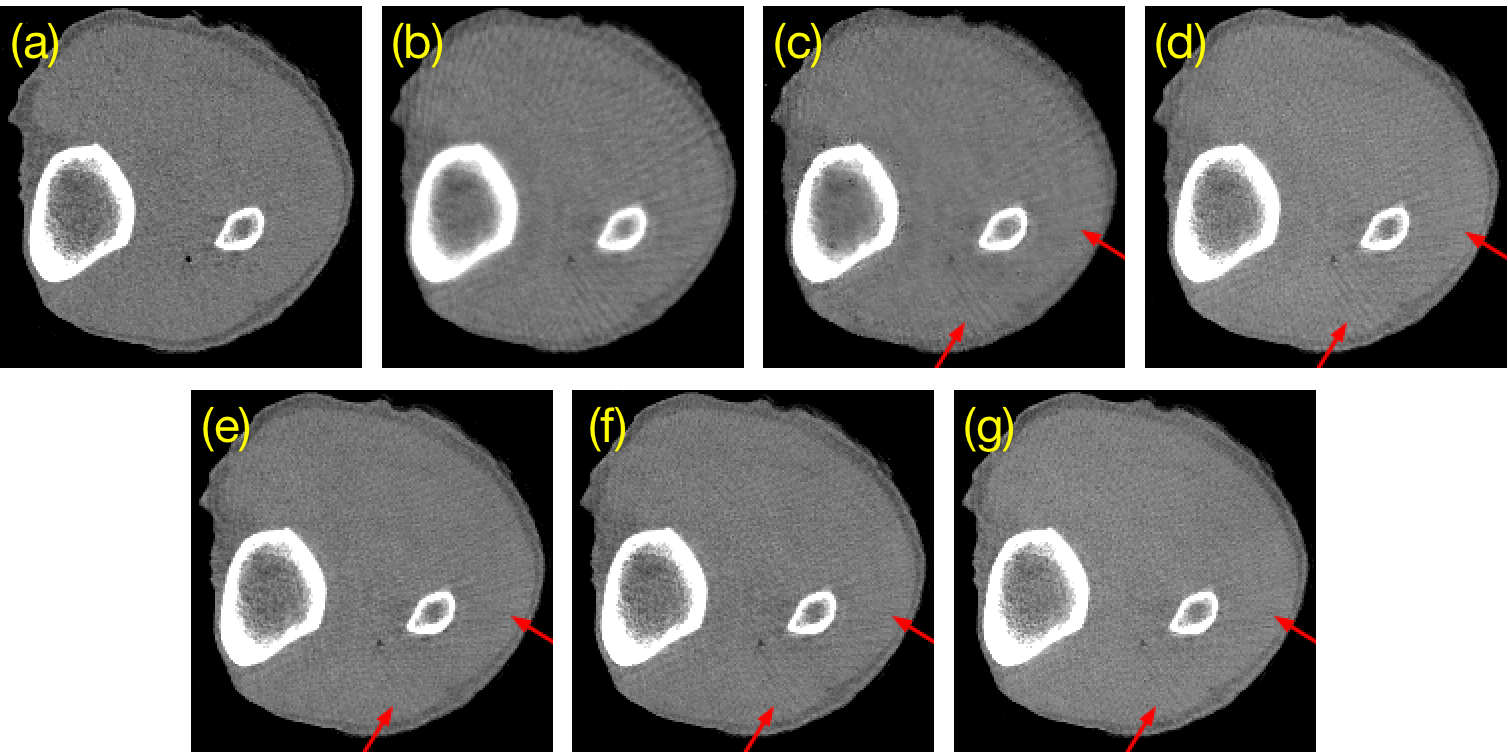}
  
    \caption{
      Qualitative sparse-view CBCT artifact reduction results by different models. The same brightness and contrast enhancement are applied to the images for better and uniform visualization. (a) $x_d$ (b) $x_s$ (c) Baseline-MSE (d) Baseline-Perceptual (e) Ours-Focus (f) Ours-FPN (g) Ours-Focus+FPN
    }
    \label{fig:visual_compare}
  \end{figure}

  \noindent \textbf{Experimental Results} Fig. \ref{fig:visual_compare} shows the qualitative results of the models. Although the baseline methods overall have some improvements over the sparse-view image, they still cannot handle the streak artifacts very well. ``Baseline-Perceptual'' produces less pronounced artifacts than ``Baseline-MSE'', which demonstrates that using perceptual loss and processing the images slice by slice in 2D give better results than MSE loss with 3D generator. Our models (Fig. \ref{fig:visual_compare} (e-f)) in general produce less artifacts than the baseline models. We can barely see the streak artifacts. 
  They generally produce similar outputs and the result from ``Ours-Focus+FPN'' is slightly better than ``Ours-FPN'' and ``Ours-Focus''. This means that using FPN as the discriminator or applying a modulation factor to the adversarial loss can indeed induce the training to artifacts reduction.
  
\begin{figure}[t]
    \centering
    \subfigure[\small Difference Maps]{
    \includegraphics[width=.225\textwidth]{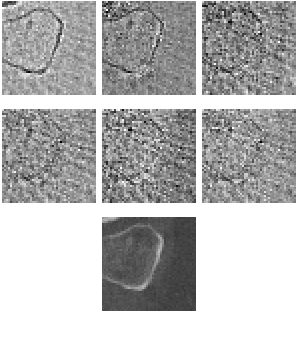}
    }
    \subfigure[Mean and Standard Deviation]{
    \includegraphics[width=.63\textwidth]{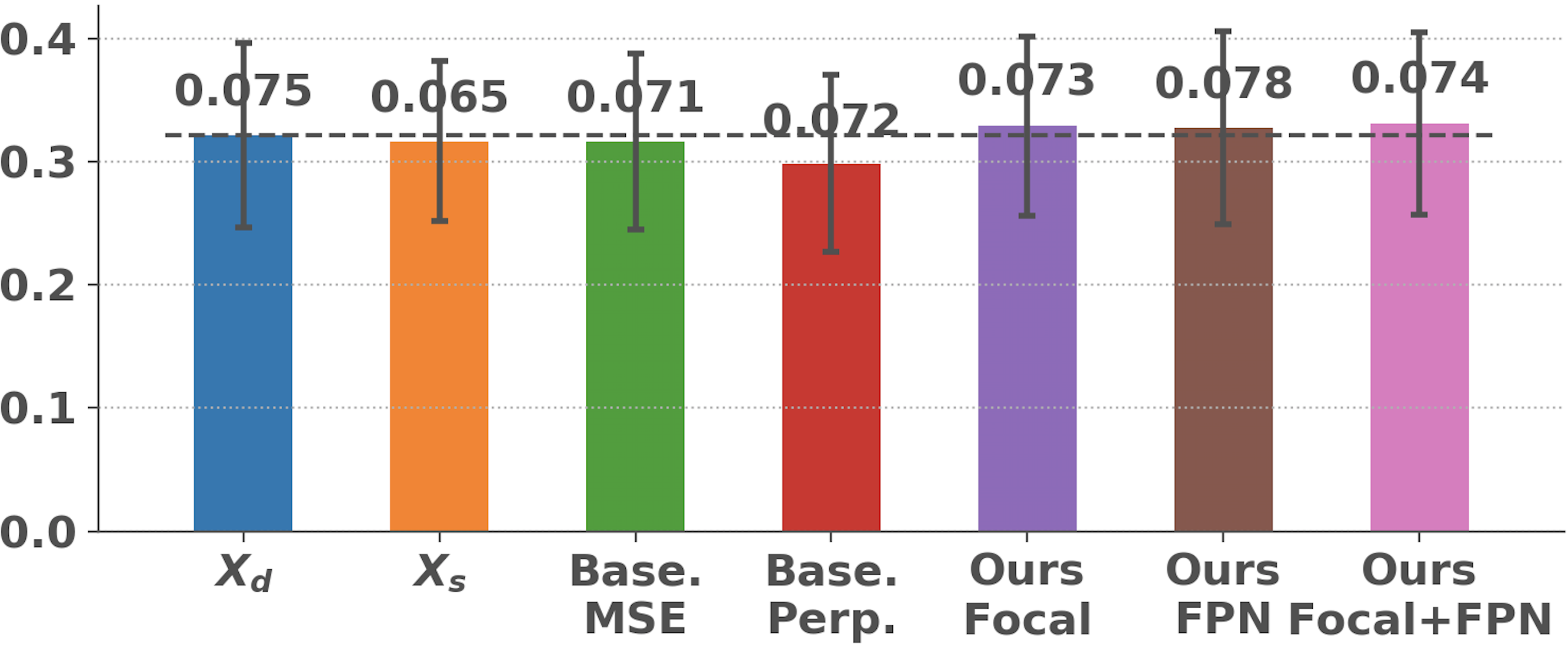}
    }
  
    \caption{
      ROI characteristics. (a) Patches are obtained by subtracting the corresponding ROI from $x_d$ (third row). First row from left to right: $x_s$, baseline-MSE, baseline-perceptual. Second row from left to right: Ours-Focus, Ours-FPN, Ours-Focus+FPN. (b) Each bar indicates the mean value of the ROI. The numbers on the top of each bar indicate the standard deviations. The vertical lines indicates the changes of the mean value when $\pm$ standard deviations is applied. Pixel values are normalized to $[0, 1]$.
    }
    \label{fig:roi_comparison}
  \end{figure}

  We further investigate the image characteristics of each model in a region of interest (ROI). A good model should have similar image characteristics to the dense-view images in ROIs. When looking at the pixel-wise difference between the dense-view ROI and the model ROI,  no structure information should be observed, resulting a random noise map. Fig. \ref{fig:roi_comparison}(a) shows the ROI differences of the models. We can see a clear bone structure from the ROI difference map between $x_s$ and $x_d$ (Fig. \ref{fig:roi_comparison}(a) third row), which demonstrates a significant difference in image characteristics between these two images. For ``Baseline-MSE'', the bone structure is less recognizable, showing more similar image characteristics. For ``Baseline-Perceptual'' and our models, we can hardly see the structural information and mostly observe random noises. This indicates that these models have very similar image characteristics to a dense-view image. We also measure the mean and standard deviation of the pixel values within the ROI. We can see that our models have very close statistics with $x_d$, especially the pixel value statistics of ``Ours-Focus'' and ``Ours-Focus+FPN'' are almost identical to $x_d$, demonstrating better image characteristics.

  We then evaluate the models quantitatively by comparing their outputs with the corresponding dense-view CBCT image. Three evaluation metrics are used: structural similarity (SSIM), peak signal-to-noise ratio (PSNR), and root mean square error (RMSE). Higher values for SSIM and PSNR and lower values for RMSE indicate better performance. We can see from Table \ref{fig:number_compare} that the baseline methods give better scores than $x_s$. Similar to the case in the qualitative evaluation, ``Baseline-Perceptual'' performs better than ``Baseline-MSE''. Our methods consistently outperform the baseline methods by a significant margin. ``Ours-FPN'' gives best performance in PSNR and RMSE. However, PSNR and RMSE only measure the pixel level difference between two images. To measure the performance in perceived similarity, SSIM is usually a better choice, and we find ``Ours-FPN+Focus'' has a slightly better performance on this metric. This confirms our observation in qualitative evaluation.
  
  \begin{table}[t]
    \centering
    \caption{Quantitative sparse-view CBCT artifact reduction results of different models.}
    \label{fig:number_compare}
    \begin{tabular}{lcx{1cm}cx{1cm}x{1cm}cx{1cm}x{1cm}x{1.5cm}}
      \toprule
                                & \phantom{a} & \multicolumn{1}{c}{\multirow{2}{*}{$x_s$}} & \phantom{a} & \multicolumn{2}{c}{\textbf{Baseline}}                       & \phantom{a} & \multicolumn{3}{c}{\textbf{Ours}}                                                            \\
                                 \cmidrule{5-6} \cmidrule{8-10}
                                &  & \multicolumn{1}{c}{}                   &  & \multicolumn{1}{c}{\textbf{MSE}} & \multicolumn{1}{c}{\textbf{Perc.}} & & \multicolumn{1}{c}{\textbf{Focus}} & \multicolumn{1}{c}{\textbf{FPN}} & \multicolumn{1}{c}{{\scriptsize \textbf{FPN+Focus}}} \\ \midrule
    \textbf{SSIM}                       &  & 0.839                                  &  & 0.849                   & 0.858                    &  & 0.879                     & 0.871                   & \textbf{0.884}                         \\
    \textbf{PSNR (dB)}                  &  & 34.07                                  &  & 34.24                   & 35.39                    &  & 36.26                     & \textbf{36.38}                   & 36.14                         \\
    \textbf{RMSE ($\mathbf{10^{-2}}$)} &  & 1.98                                   &  & 1.96                    & 1.70                     &  & 1.54                      & \textbf{1.52}                    & 1.56                         \\ \bottomrule
    \end{tabular}
    \vspace{-0.5em}
  \end{table}
  
  \section{Conclusion}
  
  We have presented a novel approach to reducing artifacts from sparsely-reconstructed CBCT images. To our best knowledge, this is the first work that addresses artifacts introduced by sparse-view CBCT reconstruction in a post-processing step. We target this problem using an image-to-image generative model with a perceptual loss as regulation. The model generates perceptually realistic  outputs while making the artifacts less pronounced. To further suppress the streak artifacts, we have also proposed a novel FPN based discriminator and a focus map to induce the adversarial training. Experimental results show that the proposed mechanism addresses the streak artifacts much better, and the proposed models outperform strong baseline methods both qualitatively and quantitatively. \vspace{0.5em}
  
  \noindent \textbf{Acknowledgement.} The work presented here was supported in part by New York State through the Goergen Institute for Data Science at the University of Rochester and the corporate sponsor Carestream Health Inc.
  
  \bibliographystyle{splncs03}
  \bibliography{references}
  
  \end{document}